\documentclass[11pt,a4paper]{article}
\usepackage[hyperref]{acl2019}
\usepackage{times}
\usepackage{latexsym}
\usepackage{graphicx}
\usepackage{verbatim}
\usepackage{soul,color}
\usepackage{multirow}
\usepackage{url}
\usepackage{mathtools}
\usepackage[]{algorithm2e}
\usepackage{bbold}

\usepackage{url}

\aclfinalcopy 
\def\aclpaperid{12} 

\title{Semi-Supervised Neural Text Generation by Joint Learning of Natural Language Generation and Natural Language Understanding Models} 

\author{Raheel Qader$^1$ \qquad
  Fran\c{c}ois Portet$^2$ \qquad
  Cyril Labb{\'e}$^2$  \\
  Univ. Grenoble Alpes, LIG\\
  38000 Grenoble, France \\
  {$^1$\tt raheel.qader@univ-grenoble-alpes.fr}\\
  {$^2$\tt \{francois.portet, cyril.labbe\}@imag.fr}}

\date{}

\begin{document}
\maketitle
\begin{abstract}
In Natural Language Generation (NLG), End-to-End (E2E) systems trained through deep learning have recently gained a strong interest. Such deep models need a large amount of carefully annotated data to reach satisfactory performance. However, acquiring such datasets for every new NLG application is a tedious and time-consuming task. In this paper, we propose a semi-supervised deep learning scheme that can learn from non-annotated data and annotated data when available. It uses an NLG and a Natural Language Understanding (NLU) sequence-to-sequence models which are learned jointly to compensate for the lack of annotation. Experiments on two benchmark datasets show that, with limited amount of annotated data, the method can achieve very competitive results while not using any pre-processing or re-scoring tricks. These findings open the way to the exploitation of non-annotated datasets which is the current bottleneck for the E2E NLG system development to new applications.
\end{abstract}

\section{Introduction}

Natural Language Generation (NLG) is an NLP task that consists in generating a sequence of natural language sentences from non-linguistic data. Traditional approaches of NLG consist in creating specific algorithms in the consensual NLG pipeline \cite{Gatt2017}, but there has been recently a strong interest in End-to-End (E2E) NLG systems which are able to jointly learn sentence planning and surface realization \cite{Dusek2016,agarwal2018char2char,JuraskaKBW18,gehrmann2018end}. Probably the most well known effort of this trend is the E2E NLG challenge \cite{NovikovaDR17} whose task was to perform sentence planing and realization from dialogue act-based Meaning Representation (MR) on \emph{unaligned} data. For instance, Figure~\ref{tab:sample_nlg} presents, on the upper part, a meaning representation and on the lower part, one possible textual realization to convey this meaning. Although the challenge was a great success, the data used in the challenge contained a lot of redundancy of structure and a limited amount of concepts and several reference texts per MR input (8.1 in average). This is an ideal case for machine learning but is it the one that is encountered in all E2E NLG real-world applications?

\begin{figure}[t]
\footnotesize{
\begin{tabular}{p{7.2cm}}
\hline
\textbf{Source sequence (MR):} \\
name{[}The Eagle{]}, eatType{[}coffee shop{]}, food{[}French{]}, priceRange{[}moderate{]}, customerRating{[}3/5{]}, area{[}riverside{]}, kidsFriendly{[}yes{]}, near{[}Burger King{]}
\\ \hline
\textbf{Target sequence (natural language):} \\ 
The three star coffee shop, The Eagle, gives families a mid-priced dining experience featuring a variety of wines and cheeses. Find The Eagle near Burger King.   \\ \hline
\end{tabular}
}
  \caption{Example of Meaning Representation (MR) and one of its paired possible text realizations. This is a excerpt of the E2E NLG challenge dataset.}
  \label{tab:sample_nlg}
\end{figure}

In this work, we are interested in learning E2E models for real world applications in which there is a low amount of annotated data. Indeed, it is well known that neural approaches need a large amount of carefully annotated data to be able to induce NLP models. For the NLG task, that means that MR and (possibly many) reference texts must be \emph{paired} together so that supervised learning is made possible. In NLG, such paired datasets are rare and remains tedious to acquire \cite{NovikovaDR17,gardent2017creating,qader_hal-01950467}. On the contrary, large amount of \emph{unpaired} meaning representations and texts can be available but cannot be exploited for supervised learning.

In order to tackle this problem, we propose a semi-supervised learning approach which is able to benefit from unpaired (non-annotated) dataset which are much easier to acquire in real life applications. In an unpaired dataset, only the input data is assumed to be representative of the task. In such case, autoencoders can be used to learn an (often more compact) internal representation of the data. Monolingual word embeddings learning also benefit from unpaired data. However, none of these techniques are fit for the task of generating from a constrained MR representation. Hence, we extend the idea of autoencoder which is to regenerate the input sequence by using an NLG and an NLU models. To learn the NLG model, the input text is fed to the NLU model which in turn feeds the NLG model. The output of the NLG model is compared to the input and a loss can be computed. A similar strategy is applied for NLU. This approach brings several advantages: 1) the learning is performed from a large unpaired (non-annotated) dataset and a small amount of paired data to constrain the inner representation of the models to respect the format of the task (here MR and abstract text); 2) the architecture is completely differentiable which enables a fully joint learning; and 3) the two NLG and NLU models remain independent and can thus be applied to different tasks separately. 

The remaining of this paper gives some background about seq2seq models (Sec~\ref{sec:background}) before introducing the joint learning approach (Sec~\ref{sec:method}). Two benchmarks, described in~Sec~\ref{sec:corpus}, have been used to evaluate the method and whose results are presented in~Sec~\ref{sec:expe}. The method is then positioned with respect to the state-of-the-art in Sec~\ref{sec:RW} before providing some concluding remarks in Sec~\ref{sec:conclu}.

\section{Background: E2E systems}\label{sec:background}

E2E Natural Language Generation systems are typically based on the Recurrent Neural Network (RNN) architecture consisting of an encoder and a decoder also known as seq2seq~\cite{sutskever2014sequence}. The encoder takes a sequence of source words $\mathbf{x}~=~\{{x_1},{x_2}, ..., {x_{T_x}}\}$ and encodes it to a fixed length vector. The decoder then decodes this vector into a sequence of target words $\mathbf{y}~=~\{{y_1},{y_2}, ..., {y_{T_y}}\}$. Seq2seq models are able to treat variable sized source and target sequences making them a great choice for NLG and NLU tasks.

More formally, in a seq2seq model, the recurrent unit of the encoder, at each time step $t$ receives an input word $x_t$ (in practice the embedding vector of the word) and a previous hidden state ${h_t-1}$ then generates a new hidden state $h_t$ using:

\begin{equation}
    {h_t} = f({h_{t-1}}, x_t),
\end{equation}

\noindent where the function $f$ is an RNN unit such as Long Short-Term Memory (LSTM)~\cite{hochreiter1997long} or Gated Recurrent Unit (GRU)~\cite{cho2014learning}. Once the encoder has treated the entire source sequence, the last hidden state ${h_{T_x}}$ is passed to the decoder. To generate the sequence of target words, the decoder also uses an RNN and computes, at each time step, a new hidden state $s_t$ from its previous hidden state $s_{t-1}$ and the previously generated word $y_{t-1}$. At training time, $y_{t-1}$ is the previous word in the target sequence (teacher-forcing). Lastly, the conditional probability of each target word $y_t$ is computed as follows:

\begin{equation}
P(y_t|\mathbf{y}_{{<}t},\mathbf{x}) = softmax(W[s_t,c_t]{+}b),
\end{equation}


\noindent where $W$ and $b$ are a trainable parameters used to map the output to the same size as the target vocabulary and $c_t$ is the context vector obtained using the sum of hidden states in the encoder, weighted by its attention~\cite{bahdanau2014neural, luong2015effective}. The context is computed as follow:
\vspace{-1mm}
\begin{equation}
c_t = \sum_{i=1}^{T_x}\alpha_{i}^{t}\,h_i
\end{equation}

Attention weights $\alpha_{i}^{t}$ are computed by applying a softmax function over a score calculated using the encoder and decoder hidden states:

\begin{equation}
\alpha_{i}^{t}=softmax(e_{i}^{t})
\end{equation}

\begin{equation}
e_{i}^{t} = score(s_t,h_i)
\end{equation}

The choice of the score adopted in this papers is based on the \textit{dot attention} mechanism introduced in~\mbox{\cite{luong2015effective}}. The attention mechanism helps the decoder to find relevant information on the encoder side based on the current decoder hidden state.
 
\section{Joint NLG/NLU learning scheme}\label{sec:method}

\begin{figure}[t]
  \includegraphics[width=0.5\textwidth]{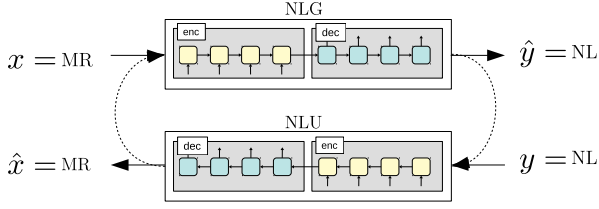}
  \caption{The joint NLG/NLU learning scheme. Dashed arrows between NLG and NLU models show data flow in the case of learning with unpaired data.}
  \label{fig:dual_learning}
\end{figure}

The joint NLG/NLU learning scheme is shown in Figure~\ref{fig:dual_learning}. It consists of two seq2seq models for NLG and NLU tasks. Both models can be trained separately on paired data. In that case, the NLG task is to predict the text $\hat{y}$ from the input MR $x$ while the NLU task is to predict the MR $\hat{x}$ from the input text $y$.  On unpaired data, the two models are connected through two different loops. In the first case, when the unpaired input source is text, $y$ is provided to the NLU models which feeds the NLG model to produce $\hat{y}$. A loss is computed between $y$ and $\hat{y}$ (but not between $\hat{x}$ and $x$ since $x$ is unknown). In the second case, when the input is only MR, $x$ is provided to the NLG model which then feeds the NLU model and finally predicts $\hat{x}$. Similarly, a loss is computed between $x$ and $\hat{x}$ (but not between $\hat{y}$ and $y$ since $y$ is unknown). This section details these four steps and how the loss is backpropagated through the loops.\\

\noindent\textbf{Learning with Paired Data:}

The NLG model is a seq2seq model with attention as described in~section~\ref{sec:background}. It takes as input a MR and generates a natural language text. The objective is to find the model parameters $\theta^{nlg}$ such that they minimize the loss which is defined as follows:

\begin{equation}
\mathcal{L}_{p}^{nlg} = -\frac{1}{T_y}\sum_{t=1}^{T_y}logP(y_t|\mathbf{x};\theta^{nlg})
\label{eq:loss_p_nlg}
\end{equation}

The NLU model is based on the same architecture but takes a natural language text and outputs a MR and its loss can be formulated as:

\begin{equation}
\mathcal{L}_{p}^{nlu} = -\frac{1}{T_x}\sum_{t=1}^{T_x}logP(x_t|\mathbf{y};\theta^{nlu})
\label{eq:loss_p_nlu}
\end{equation}\\

\noindent\textbf{Learning with Unpaired Data:}

When data are unpaired, there is also a loop connection between the two seq2seq models. This is achieved by feeding MR to the NLG model in order to generate a sequence of natural language text $\hat y$ by applying an argmax over the probability distribution at each time step ($\hat y_t = \mbox{argmax}P(y_t|\mathbf{x};\theta^{nlg})$).
This text is then fed back into the NLU model which in turn generates an MR. Finally, we compute the loss between the original MR and the reconstructed MR:

\begin{equation}
\mathcal{L}_{u}^{nlu} = -\frac{1}{T_x}\sum_{t=1}^{T_x}logP(x_t|\mathbf{x}; \theta^{nlg}, \theta^{nlu})
\label{eq:loss_u_nlu}
\end{equation}

The same can be applied in the opposite direction where we feed text to the NLU model and then the NLG model reconstructs back the text. This loss is given by:

\begin{equation}
\mathcal{L}_{u}^{nlg} = -\frac{1}{T_y}\sum_{t=1}^{T_y}logP(y_t|\mathbf{y}; \theta^{nlg}, \theta^{nlu})
\label{eq:loss_u_nlg}
\end{equation}

To perform joint learning, all four losses are summed together to provide the uniq loss $\mathcal{L}$ as follows:

\begin{equation}
\mathcal{L} = \alpha{\cdot}\mathcal{L}_{p}^{nlg} + \beta{\cdot}\mathcal{L}_{p}^{nlu} + \gamma{\cdot}\mathcal{L}_{u}^{nlg} + \delta{\cdot}\mathcal{L}_{u}^{nlu}
\end{equation}

The weights $\alpha, \beta, \delta$ and $\gamma \in [0,1]$ are defined to fine tune the contribution of each task and data to the learning or to bias the learning towards one specific task. We show in the experiment section the impact of different settings. 


Since the loss functions in~Equation~\ref{eq:loss_p_nlg}~and~\ref{eq:loss_p_nlu} force the model to generate a sequence of words based on the target and the losses in~Equation~\ref{eq:loss_u_nlg}~and~\ref{eq:loss_u_nlu} force the model to reconstruct back the input sequence, this way the model is encouraged to generate text that is supported by the facts found in the input sequence. It is important to note that the gradients based on $\mathcal{L}_{p}^{nlg}$ and $\mathcal{L}_{p}^{nlu}$ can only backpropagate through their respective model (i.e., NLG and NLU), while $\mathcal{L}_{u}^{nlg}$ and $\mathcal{L}_{u}^{nlu}$ gradients should backpropagate through both models.\\

\noindent\textbf{Straight-Through Gumbel-Softmax:}

A major problem with the proposed joint learning architecture in the unpaired case  is that the model is not fully differentiable. Indeed, given the input $x$ and the intermediate output $\hat{y}$, the $\mathcal{L}_{u}^{nlu}$ and the NLG parameter $\theta_{nlg}$, the gradient is computed as: 
\begin{equation}
 \frac{\partial \mathcal{L}_{u}^{nlu}}{\partial \theta_{nlg}}
   = \sum_t^T \left( \frac{\partial \mathcal{L}_{u}^{nlu}}{\partial \hat{y}_t}
      + \frac{\partial \hat{y}_t}{\partial p_{y_t}}
     + \frac{\partial p_{y_t}}{\partial \theta_{nlg}} 
 \right) 
\end{equation}

At each time step $t$, the output probability $p_{y_t}$ is computed trough the softmax layer and $\hat{y}_t$ is obtained using $\hat{y}_t = onehot(argmax_w p_{y_t}[w])$ that is the word index $w$ with maximum probability at time step $t$. 
To address this problem, one solution is to replace this operation by the identity matrix $\frac{\partial \hat{y}_t}{\partial p_{y_t}} \approx \mathbb{1}$. This approach is called the Straight-Through (ST) estimator, which simply consists of backpropagating through the argmax function as if it had been the identity function \cite{BengioLC13,yin2018understanding}. 

A more principled way of dealing with the non-differential nature of argmax, is to use the Gumbel-Softmax which proposes a continuous approximation to sampling from a categorical distribution~\cite{jang2016categorical}. Hence, the discontinuous argmax is replaced by a differentiable and smooth function. More formally, consider a $k$-dimensional categorical distribution $u$ with probabilities $\pi_1, \pi_2, ..., \pi_k$. Samples from $u$ can be approximated using:

\begin{equation}
y_i = \frac{\text{exp}((\log(\pi_i)+g_i)/\tau)}{\sum_{j=1}^k \text{exp}((\log(\pi_j)+g_j)/\tau)}
\label{eq:gumbel}
\end{equation}

\begin{equation}
gi = -\mbox{log}(-\mbox{log}(u_i))
\end{equation}
\begin{equation}
u_i \sim \mbox{Uniform}(0, 1),
\end{equation}

\noindent where $g_i$ is the Gumbel noise drawn from a uniform distribution and $\tau$ is a temperature parameter. The sample distribution from the Gumbel-Softmax resembles the argmax operation as $\tau \rightarrow 0$, and it becomes uniform when $\tau \rightarrow \infty$. 

Although Gumbel-Softmax is differentiable, the samples drawn from it are not adequate input to the subsequent models which expect a discrete values in order to retrieve the embedding matrix of the input words. So, instead, we use the Straight-Through (ST) Gumbel-Softmax which is basically the discrete version of the Gumbel-Softmax. During the forward phase, ST Gumbel-Softmax discretizes $y$~in~Equation~\ref{eq:gumbel} but it uses the continuous approximation in the backward pass. Although the Gumbel-Softmax estimator is biased due to the sample mismatch between the backward and forward phases, many studies have shown that ST Gumbel-Softmax can lead to significant improvements in several tasks~\cite{choi2018learning, gu2018neural, tjandra2018end}.

\section{Dataset}\label{sec:corpus}
The models developed were evaluated on two datasets. The first one is the E2E NLG challenge dataset \cite{NovikovaDR17} which contains 51k of annotated samples. The second one is the Wikipedia Company Dataset \cite{qader_hal-01950467} which consists of around 51K of noisy MR-abstract pairs of company descriptions. 

\subsection{E2E NLG challenge Dataset}

The E2E NLG challenge Dataset has become one of the benchmarks of reference for end-to-end sentence-planning NLG systems. It is still one of the largest dataset available for this task. The dataset was collected 
via crowd-sourcing using pictorial representations in the domain of restaurant recommendation.

Although the E2E challenge dataset contains more than 50k samples, each MR is associated on average with 8.1 different reference utterances leading to around 6K unique MRs. Each MR consists of 3 to 8 slots, such as \emph{name}, \emph{food} or \emph{area}, and their values and slot types are fairly equally distributed. The majority of MRs consist of 5 or 6 slots while human utterances consist mainly of one or two sentences only. The vocabulary size of the dataset is of 2780 distinct tokens.

\subsection{The Wikipedia Company Dataset}\label{sec:wikipedia_dataset}
The wikipedia company dataset \cite{qader_hal-01950467}, is composed of a  set of company data from English Wikipedia. The dataset contains 51k samples where each sample is composed of up to 3 components: the Wikipedia article abstract, the Wikipedia article body, and the infobox which is a set of attribute--value pairs containing primary information about the company (\emph{founder}, \emph{creation date} etc.). The infobox part was taken as MR where each attribute--value pair was represented as a sequence of string \texttt{attribute [value]}. 
The MR representation is composed of 41 attributes with 4.5 attributes per article and 2 words per value in average. The abstract length is between 1 to 5 sentences. The vocabulary size is of 158464 words.

The Wikipedia company dataset contains much more lexical variation and semantic information than the E2E challenge dataset. Furthermore, company texts have been written by humans within the Wikipedia ecosystem and not during a controlled experiment whose human engagement was unknown. Hence, the Wikipedia dataset seems an ecological target for research in NLG. However, as pointed out by the authors, the Wikipedia dataset is not ideal for machine learning. First, the data is not controlled and each article contains only one reference (vs. 8.1 for the E2E challenge dataset). Second the abstract, the body and the infobox are only loosely correlated. Indeed, the meaning representation coverage is poor since, for some MR, none of the information is found in the text and vice-versa. To give a rough estimate of this coverage, we performed an analysis of 100 articles randomly selected in the test set. Over 868 total slot instances, 28\% of the slots in the infobox cannot be found in their respective abstract text, while 13\% are missing in the infobox. 

Despite these problems, we believe the E2E and the Wikipedia company datasets can provide contrasted evaluation, the first being well controlled and lexically focused, the latter representing the kind of data that can be found in real situations and that E2E systems must deal with in order to percolate in the society.

\section{Experiments}\label{sec:expe}
The performance of the joint learning architecture was evaluated on the two datasets described in the previous section. The joint learning model requires a paired and an unpaired dataset,
so each of the two datasets was split into several parts. 

\textbf{E2E NLG challenge Dataset:}
The training set of the E2E challenge dataset which consists of 42K samples was partitioned into a 10K paired and 32K unpaired datasets by a random process. The unpaired database was composed of two sets, one containing MRs only and the other containing natural texts only. This process resulted in 3 training sets: paired set, unpaired text set and unpaired MR set. The original development set (4.7K) and test set (4.7K) of the E2E dataset have been kept.

\textbf{The Wikipedia Company Dataset:}
The Wikipedia company dataset presented in Section~\ref{sec:wikipedia_dataset} was filtered to contain only companies having abstracts of at least 7 words and at most 105 words. As a result of this process, 43K companies were retained. The dataset was then divided into: a training set (35K), a development set (4.3K) and a test set (4.3K).
Of course, there was no intersection between these sets. 

The training set was also partitioned in order to obtain the paired and unpaired datasets. 
Because of the loose correlation between the MRs and their corresponding text, the paired dataset was selected such that it contained the infobox values with the highest similarity with its reference text. The similarity was computed using ``difflib'' library\footnote{\url{https://docs.python.org/2/library/difflib.html}}, which is an extension of the Ratcliff and Obershelp algorithm \citep{ratcliff1988pattern}. 
The paired set was selected in this way (rather than randomly) to get samples as close as possible to a carefully annotated set.  At the end of partitioning, the following training sets were obtained: paired set (10.5K), unpaired text set (24.5K) and unpaired MR set (24.5K). \\

The way the datasets are split into paired and unpaired sets is artificial and might be biased particularly for the E2E dataset as it is a rather easy dataset. This is why we included the Wikipedia dataset in our study since the possibility of having such bias is low because 1) each company summary/infobox was written by different authors at different time within the wikipedia eco-system making this data far more natural than in the E2E challenge case, 2) there is a large amount of variation in the dataset, and 3) the dataset was split in such a way that the paired set contains perfect matches between the MR and the text, while reserving the least matching samples for the the unpaired set (i.e., the more representative of real-life Wikipedia articles). As a result, the paired and unpaired sets of the Wikipedia dataset are different from each other and the text and MR unpaired samples are only loosely correlated.

\subsection{Evaluation with Automatic Metrics}

\begin{table*}[t]
\begin{tabular}{p{2cm}lllllll|lll}
\cline{6-11}
                                                  &                               &                              &                               &          & \multicolumn{3}{c}{NLG} & \multicolumn{3}{c}{NLU}     \\ \hline

\multicolumn{1}{l}{System}                      & \multicolumn{1}{l}{$\alpha$}    & \multicolumn{1}{l}{$\beta$}    & \multicolumn{1}{l}{$\gamma$}   & $\delta$ & BLEU   & Rouge-L  & Meteor & Precision & Recall & F-score \\ \hline
\multicolumn{1}{l}{Paired}                      & \multicolumn{1}{l}{-}    & \multicolumn{1}{l}{-}    & \multicolumn{1}{l}{-}   & - & 0.60  & 0.64   & 0.42    & 0.74      & {\bf 0.83}   & 0.78    \\ \hline
\multicolumn{1}{l}{\multirow{4}{0pt}{Paired + Unpaired}} & \multicolumn{1}{l}{0.25} & \multicolumn{1}{l}{0.25} & \multicolumn{1}{l}{1}   & 1 & {\bf 0.64}$^\dagger$   & 0.66$^\dagger$   & 0.43    & 0.73      & 0.78   & 0.76    \\ \cline{2-11}
\multicolumn{1}{l}{}                            & \multicolumn{1}{l}{0.1}  & \multicolumn{1}{l}{0.1}  & \multicolumn{1}{l}{1}   & 1 &  {\bf 0.64}$^\dagger$  & {\bf 0.67}$^\dagger$  & 0.42    & 0.73      & 0.74   & 0.74    \\ \cline{2-11} 
\multicolumn{1}{l}{}                            & \multicolumn{1}{l}{1}  & \multicolumn{1}{l}{0.1}    & \multicolumn{1}{l}{1}   & 1 & 0.63$^\dagger$  & {\bf 0.67}$^\dagger$   & 0.43$^\dagger$    & 0.72      & 0.78   & 0.75    \\ \cline{2-11} 
\multicolumn{1}{l}{}                            & \multicolumn{1}{l}{1}  & \multicolumn{1}{l}{0.1}    & \multicolumn{1}{l}{1} & 0.1 & {\bf 0.64}$^\dagger$  & {\bf 0.67}$^\dagger$   & {\bf 0.45}$^\dagger$    & {\bf 0.77}     & {\bf 0.83}   & {\bf 0.80}    \\ \hline
\end{tabular}
\caption {Results on the test set of E2E dataset. $^\dagger$ indicates t-test $p<0.001$ against the paired NLG results.} \label{tab:e2e_results} 
\end{table*}

\begin{table*}[t]
\begin{tabular}{p{2cm}lllllll|lll}
\cline{6-11}
                                                  &                               &                              &                               &          & \multicolumn{3}{c}{NLG} & \multicolumn{3}{c}{NLU}     \\ \hline

\multicolumn{1}{l}{System}                      & \multicolumn{1}{l}{$\alpha$}    & \multicolumn{1}{l}{$\beta$}    & \multicolumn{1}{l}{$\gamma$}   & $\delta$ & BLEU   & Rouge-L  & Meteor & Precision & Recall & F-score \\ \hline
\multicolumn{1}{l}{Paired~~~~~~}              & \multicolumn{1}{l}{-}        & \multicolumn{1}{l}{-}       & \multicolumn{1}{l}{-}        & -        & 0.08  & 0.24   & 0.11    & {\bf 0.20}      & 0.33   & 0.25 \\ \hline
\multicolumn{1}{l}{\multirow{4}{1pt}{Paired + Unpaired}} & \multicolumn{1}{l}{0.25}     & \multicolumn{1}{l}{0.25}    & \multicolumn{1}{l}{1}        & 1        & 0.02$^\dagger$  & 0.15$^\dagger$   & 0.07$^\dagger$    & {\bf 0.20}      & {\bf 0.43}   & {\bf 0.27}   \\ \cline{2-11} 
\multicolumn{1}{l}{}                           & \multicolumn{1}{l}{0.1}      & \multicolumn{1}{l}{0.1}     & \multicolumn{1}{l}{1}        & 1        & 0.04$^\dagger$  & 0.18$^\dagger$   & 0.08$^\dagger$    & 0.08      & 0.22   & 0.12   \\ \cline{2-11} 
\multicolumn{1}{l}{}                           & \multicolumn{1}{l}{1}      & \multicolumn{1}{l}{0.1}       & \multicolumn{1}{l}{1}        & 1        & 0.08  & {\bf 0.26}$^\dagger$   & {\bf 0.12}$^\dagger$    & 0.18      & 0.42   & 0.25  \\ \cline{2-11} 
\multicolumn{1}{l}{}                           & \multicolumn{1}{l}{1}      & \multicolumn{1}{l}{0.1}       & \multicolumn{1}{l}{1}      & 0.1        & {\bf 0.09}$^\dagger$  & {\bf 0.26}$^\dagger$   & {\bf 0.12}$^\dagger$    & {\bf 0.20}      & 0.35   & 0.26    \\ \hline
\end{tabular}
\caption {Results on the test set of Wikipedia company dataset. $^\dagger$ indicates t-test $p<0.001$ against the Paired NLG results.} \label{tab:wiki_results} 
\end{table*}

For the experiments, each seq2seq model was composed of 2 layers of Bi-LSTM in the encoder and two layers of LSTM in the decoder with 256 hidden units and \textit{dot attention} trained using Adam optimization with learning rate of 0.001. The embeddings had 500 dimensions and the vocabulary was limited to 50K words. The Gumbel-Softmax temperature $\tau$ was set to 1. Hyper-parameters tuning was performed on the development set and models were trained until the loss on the development set stops decreasing for several consecutive iterations. All models were implemented with PyTorch library.

Results of the experiment on the E2E challenge data are summarized Table~\ref{tab:e2e_results} for both the NLG and the NLU tasks. BLEU, Rouge-L and Meteor were computed using the E2E challenge metrics script\footnote{\url{https://github.com/tuetschek/e2e-metrics}} with default settings. NLU performances were computed at the slot level. 
The model learned using paired+unpaired methods shows significant superior performances than the paired version. Among the paired+unpaired methods, the one of last row exhibits the highest balanced score between NLG and NLU. This is achieved when the weights $\alpha$ and $\gamma$ favor the NLG task against NLU ($\beta=\delta=0.1$). This setting has been chosen since the NLU task converged much quicker than the NLG task. Hence lower weight for NLU during the learning avoided over-fitting. This best system exhibits similar performances than the E2E challenge winner for ROUGE-L and METEOR whereas it did not use any pre-processing (delexicalisation, slot alignment, data augmentation) or re-scoring and was trained on far less annotated data.

\begin{table*}[!htb]
\small
\begin{tabular}{ll}
\hline

\multicolumn{1}{l}{Input}                  & \multicolumn{1}{p{12.3cm}}{name{[} the punter {]}, eattype{[} restaurant {]}, food{[} indian {]}, pricerange{[} moderate {]}, customer\_rating{[} 1 out of 5 {]}, area{[} city centre {]}, familyfriendly{[} no {]}, near{[} express by holiday inn {]}} \\ \hline
\multicolumn{1}{l}{Reference}              & \multicolumn{1}{p{11cm}}{the punter is a restaurant providing indian food in the moderate price range. it is located in the city centre. it is near express by holiday inn. its customer rating is 1 out of 5.}                                        \\ \hline
\multicolumn{1}{l}{Paired model}          & \multicolumn{1}{p{11cm}}{the punter is a moderately priced indian restaurant in the city centre near express by holiday inn. it has a customer rating of 1 out of 5.}                                                                                  \\ \hline
\multicolumn{1}{l}{Paired+unpaired model} & \multicolumn{1}{p{11cm}}{the punter is a restaurant providing indian food in the moderate price range. it is located in the city centre. it is near express by holiday inn. its customer rating is 1 out of 5.} \\ \hline

\hline \hline
\multicolumn{1}{l}{Input}                  & \multicolumn{1}{p{11cm}}{name{[} the cricketers {]}, eattype{[} restaurant {]}, food{[} chinese {]}, pricerange{[} less than £20 {]}, customer\_rating{[} low {]}, area{[} city centre {]}, familyfriendly{[} no {]}, near{[} all bar one {]}}         \\ \hline
\multicolumn{1}{l}{Reference}              & \multicolumn{1}{p{11cm}}{the cricketers is a restaurant providing chinese for under £20. it has a low customer rating. it is located in the city center. it is not family friendly. it is near all bar one.}                                           \\ \hline
\multicolumn{1}{l}{Paired model}          & \multicolumn{1}{p{11cm}}{the cricketers is a chinese restaurant located in the city centre near all bar one that is not family friendly and has a low customer rating and a price range of less than £20.}                                             \\ \hline
\multicolumn{1}{l}{Paired+unpaired model} & \multicolumn{1}{p{11cm}}{the cricketers is a restaurant providing chinese food in the less than £20 price range. it is located in the city centre. it is near all bar one. its customer rating is low.} \\ \hline

\hline \hline

\multicolumn{1}{l}{Input}           & \multicolumn{1}{p{9.7cm}}{name1{[} sedgwick group {]}, headquarters1{[} london {]}, founded1{[} 1998 {]}, industry1{[} insurance {]}, key\_people1{[} sax riley (chairman) {]}, key\_people2{[} rob whitecooper (ceo) {]}, defunct1{[} 1998 {]}}                                                                                                                                                                                                                            \\ \hline

\multicolumn{1}{l}{Reference}       & \multicolumn{1}{p{11cm}}{sedgwick group plc was a very large british insurance broker . it was listed on the london stock exchange and was a constituent of the ftse 100 index . however it was acquired by marsh \& mclennan in 1998 .}                                                                                                                                                                                                                                   \\ \hline
\multicolumn{1}{l}{Paired model}          & \multicolumn{1}{p{11cm}}{the porton group was a british insurance company . it was founded in 1998 by former mckinsey and entrepreneur erik muir . it was the first property manager until it was acquired by the saatchi group in 1998 .}                                                                                                                                                                                                                                 \\ \hline
\multicolumn{1}{l}{Paired+unpaired model} & \multicolumn{1}{p{11cm}}{sedgwick group plc was a british insurance company based in london . it was listed on the london stock exchange and was once a constituent of the ftse 100 index .}                                                                                                                                                                                                                                                                               \\ \hline

\hline \hline
\multicolumn{1}{l}{Input}           & \multicolumn{1}{p{11cm}}{name1{[} federal realty investment trust {]}, founded1{[} 1962 {]}, industry1{[} real estate investment trust {]}, key\_people1{[} joseph s vassalluzzo chairman {]}, key\_people2{[} donald c wood ceo president {]}, key\_people3{[} daniel guglielmone cfo {]}, products1{[} shopping centers {]}, founder1{[} samuel j gorlitz {]}}                                                                                                           \\ \hline
\multicolumn{1}{l}{Reference}       & \multicolumn{1}{p{11cm}}{federal realty investment trust is a real estate investment trust that invests in shopping centers in the northeastern united states , the mid-atlantic states , california , and south florida .}                                                                                                                                                                                                                                                \\ \hline
\multicolumn{1}{l}{Paired model}          & \multicolumn{1}{p{11cm}}{city capital trust trust , inc. is a real estate investment trust and investment trust . it was founded in 1962 by robert c. smith , and is based in new york city , and is headquartered in cleveland , connecticut . the company is headquartered in cleveland , florida , and has offices in new york city , new york , and geneva .}                                                                                                          \\ \hline
\multicolumn{1}{l}{Paired+unpaired model} & \multicolumn{1}{p{11cm}}{the federal realty investment trust , is a real estate investment trust that invests in shopping centers in the united states . it was founded in 1962 by robert duncan , jr. and john epstein .}  \\ \hline

\end{tabular}
\caption {Sample of generated text from the E2E and Wikipedia test sets using our systems along with the reference text.} \label{tab:sample_outputs} 
\end{table*}

Results of the experiment on Wikipedia company dataset are summarized Table~\ref{tab:wiki_results} for both the NLG and the NLU tasks. Due to noise in the dataset and the fact that only one reference is available for each sample, the automatic metrics show very low scores. This is in line with \cite{qader_hal-01950467} for which the best system obtained BLEU$=0.0413$, ROUGE-L$=0.266$ and METEOR$=0.1076$. Contrary to the previous results, the paired method brings one of the best performance
. However, the best performing system is the one of the last row which again put more emphasis on the NLG task than on the NLU one. Once again, this system obtained performances comparable to the best system of \cite{qader_hal-01950467} but without using any pointer generator or coverage mechanisms.

In order to further analyze the results, in Table~\ref{tab:sample_outputs} we show samples of the generated text by different models alongside the reference texts. The first two examples are from the model trained on the E2E NLG dataset and the last two are from the Wikipedia dataset. Although on the E2E dataset the outputs of paired and paired+unpaired models seem very similar, the latter resembles the reference slightly more and because of this it achieves a higher score in the automatic metrics. This resemblance to the reference could be attributed to the fact that we use a reconstruction loss which forces the model to generate text that is only supported by facts found in the input. As for the Wikipedia dataset examples, we can see that the model with paired+unpaired data is less noisy and the outputs are generally shorter. The model with only paired data generates unnecessarily longer text with lots of unsupported facts and repetitions. Needless to say that both models are doing lots of mistakes and this is because of all the noise contained in the training data.

\subsection{Human Evaluation}

\begin{table}[t]
\begin{small}
\centering
\begin{tabular}{lcccc}
\hline
          &cover. &non-redun.&semant.&gramm.\\
\hline
reference & 3.42& 4.25&4.19&4.13\\
\hline
paired    &2.26&{\bf 3.67}&3.28&{\bf 4.11}\\
unpaired  &{\bf 2.87}$^\dagger$& 3.63&{\bf3.67}&3.96\\
\hline
\end{tabular}
\end{small}
\caption{Results of the human evaluation per system on the Wikipedia corpus using the best unpaired system. $^\dagger$ indicates wilcoxon $p<0.05$ against the paired results. \label{tab:Human_results}}
\end{table}

 It is well know that automatic metrics in NLG are poorly predictive of human ratings although they are useful for system analysis and development \cite{Novikova2017,Gatt2017}. Hence, to gain more insight about the generation properties of each model, a human evaluation with 16 human subjects was performed on the Wikipedia dataset models. We set up a web-based experiment and used the same 4 questions as in \cite{qader_hal-01950467} which were asked on a 5-point Lickert scale: How do you judge the Information Coverage of the company summary?  How do you judge the Non-Redundancy of Information in the company summary? How do you judge the Semantic Adequacy of the company summary? How do you judge the Grammatical Correctness of the company summary?

For this experiment, 40 company summaries were selected randomly from the test set. Each participant had to treat 10 summaries by first reading the summary and the infobox, then answering the aforementioned four questions.

Results of the human experiment are reported in~Table~\ref{tab:Human_results}. The first line reports the results of the reference (i.e., the Wikipedia abstract) for comparison, while the second line is the model with paired data, and the last line is the model trained on paired+unpaired data with parameters reported in the last row of Table~\ref{tab:wiki_results}, i.e., $\alpha=\gamma=1$ and $\beta=\delta=0.1$ . It is clear from the coverage metric that no system nor the reference was seen as doing a good job at conveying the information present in the infobox. This is in line with the corpus analysis of section~\ref{sec:corpus}. However, between the automatic methods, the unpaired models exhibit a clear superiority in coverage and in semantic adequacy, two measures that are linked. On the other side, the model learned with paired data is slightly more performing in term of non-redundancy and grammaticality. The results of the unpaired model with coverage and grammaticality are equivalent to best models of~\citet{qader_hal-01950467} but for non-redundancy and semantic adequacy the result are slightly below. This is probably because the authors have used a pointer generator mechanism \cite{See2017}, a trick we avoided and which is subject of further work. 

\begin{figure}[t]
  \includegraphics[width=0.49\textwidth]{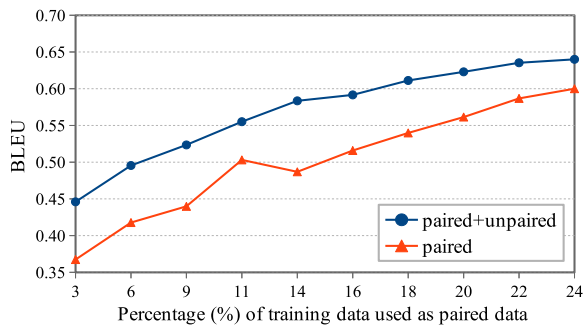}
  \caption{BLEU score as a function of percentage of paired data in the training set on the E2E dataset.}
  \label{fig:ablation_pair_percentage}
\end{figure}

These results express the difference between the learning methods: on the one hand, the unpaired learning relaxes the intermediate labels which are noisy so that the model learns to express what is really in the input (this explain the higher result for coverage) while, on the other hand, the paired learning is only constrained by the output text (not also with the NLU loss as in the unpaired case) which results in slightly more grammatical sentence to the expense of semantic coverage.

\subsection{Ablation Study}

\begin{table}[t]
\begin{tabular}{lllllll}
\hline
$\alpha$ & $\beta$   & $\gamma$ & $\delta$   & BLEU & Rouge-L & Meteor \\ \hline
1 & 0.1   & 1 & 0.1 &  {\bf 0.64}  & {\bf 0.67}   & {\bf 0.45}  \\ \hline
\hline
0 & 0.1   & 1 & 0.1 &   0.62   &    0.66     &   0.42    \\ \hline
1 & 0 & 1 & 0.1 &   0.63   &    0.67     &   0.42    \\ \hline
1 & 0.1 & 0 & 0.1   &   0.50   &    0.58     &   0.36    \\ \hline
1 & 0.1 & 1 & 0 &   0.63   &    0.66     &   0.44    \\ \hline
\end{tabular}
\caption {Effect of loss weights on the performance of the NLG model on the E2E dataset.} \label{tab:ablation_nlg} 
\end{table}

\begin{table}[t]
\begin{tabular}{lllllll}
\hline
$\alpha$ & $\beta$   & $\gamma$ & $\delta$   & Precision & Recall & F-score \\ \hline
1 & 0.1   & 1 & 0.1 &  {\bf 0.77}      & {\bf 0.83}   & {\bf 0.80}   \\ \hline
\hline
0 & 0.1   & 1 & 0.1 &   0.74   &    0.79     &   0.76    \\ \hline
1 & 0 & 1 & 0.1 &   0.74   &    0.71     &   0.73    \\ \hline
1 & 0.1 & 0 & 0.1   &   0.68   &    0.73     &   0.70    \\ \hline
1 & 0.1 & 1 & 0 &   0.75   &    0.73     &   0.74    \\ \hline
\end{tabular}
\caption {Effect of loss weights on the performance of the NLU model on the E2E dataset.} \label{tab:ablation_nlu} 
\end{table}

In this section, we further discuss different aspects of the proposed joint learning approach. In particular we are interested in studying the impact of: 1) having different amounts of paired data and 2) the weight of each loss function on the overall performance. Since only the E2E dataset is non-noisy and hence provide meaningful automatic metrics, the ablation study was performed only on this dataset.

To evaluate the dependence on the amount of paired data, the best model was re-trained by changing the size of the paired data ranging from 3\% of the training data (i.e., 1K) up to 24\% (i.e., 10K). The results are shown in~Figure~\ref{fig:ablation_pair_percentage}. The figure reveals that regardless of the amount of paired data, the joint learning approach: 1) always improves over the model with only paired data and 2) is always able to benefit from supplementary paired data. This is particularly true when the amount of paired data is very small and the difference seems to get smaller as the percentage of the paired data increases.

Next, to evaluate which of the four losses contribute most to the overall performance, the best model was re-trained in different settings. In short, in each setting, one of the weights was set to zero while the others three weights were kept similar as in the best case. The results are presented in~Table~\ref{tab:ablation_nlg} and Table~\ref{tab:ablation_nlu} for NLG and NLU tasks respectively. In these table the first line if the best model as reported in Table~\ref{tab:e2e_results}. It can be seen that all the four losses are important since setting any of the weights to zero leads to a decrease in performance. However, the results of both tables show that the most important loss is the NLG unpaired loss $\mathcal{L}_{u}^{nlg}$ since setting $\gamma$ to zeros leads to a significant reduction in the performance for both NLU and NLG.

\section{Related Work} \label{sec:RW}

The approach of joint learning has been tested in the literature in other domains than NLG/NLU for tasks such machine translation~\cite{cheng2016semi,he2016dual,tu2017neural} and speech processing~\cite{tjandra2017listening,tjandra2018end,liu2018improving}. In \cite{tu2017neural} an encoder-decoder-reconstructor for MT is proposed. The reconstructor, integrated to the NMT model, rebuilds the source sentence from the hidden layer of the output target sentence, to ensure that the information in the source side is transformed to the target side as much as possible. In \cite{tjandra2018end}, a joint learning architecture of Automatic Speech Recognition (ASR) and Text-To-Speech (TTS) is proposed which leverages unannotated data. In the unannotated case, during the learning, ASR output is fed to the TTS and the TTS output is compared with the original ASR signal input to compute a loss which is back-propagated through both modules. Regarding NLU, joint learning of NLU with other tasks remain scarce. In \citep{Yang2017}, an NLU model is jointly learned with a system action prediction (SAP) model on supervised dialogue data. The NLU model is integrated into the sequence-to-sequence SAP model so that three losses (intent prediction, slot prediction and action prediction) are used to backpropagate through both models. The paper shows that this approach is competitive against the baselines.

To the best of our knowledge, the idea of joint NLG/NLU learning has not been tested previously in NLG. In NLG E2E models \cite{Dusek2016,JuraskaKBW18}, some approaches have learned a concept extractor (which is close to but simpler than an NLU model), but this was not integrated in the NLG learning scheme and only used for output re-scoring. Probably the closest work to our is \cite{chisholm2017learning} in which a seq2seq auto-encoder was used to generate biographies from MR. In this work, the generated text of the `forward' seq2seq model was constrained by a `backward' seq2seq model, which shared parameters. However, this works differs from ours since their model was not completely differentiable. Furthermore, their NLU backward model was only used as a support for the forward NLG. Finally, the shared parameters, although in line with the definition of an auto-encoder, make each model impossible to specialize.  

\section{Conclusion and Further Work} \label{sec:conclu}
In this paper, we describe a learning scheme which provides the ability to jointly learn two models for NLG and for NLU using large amount of unannotated data and small amount of annotated data. The results obtained with this method on the E2E challenge benchmark, show that the method can achieve a similar score of the winner of the challenge \cite{JuraskaKBW18} but with far less annotated data and without using any pre-processing (delexicalisation, data augmentation) or re-scoring tricks. 
Results on the challenging Wikipedia company dataset shows that highest score can be achieve by mixing paired and unpaired datasets. These results are at the state-of-the-art level \cite{qader_hal-01950467} but without using any pointer generator or coverage mechanisms. These findings open the way to the exploitation of unannotated data since the lack of large annotated data source is the current bottleneck of E2E NLG systems development for new applications.

Next steps of the research include, replacing the ST Gumbel-Softmax with reinforcement learning techniques such as policy gradient. This is particularly interesting as with policy gradient we will be able do design reward functions that better suit the problem we are trying to solve. Furthermore,
it would be interesting to evaluate how pointer generator mechanism \cite{See2017} and coverage mechanism \cite{Tu2016} can be integrated in the learning scheme to increase the non-redundancy and coverage performance of the generation.

\section*{Acknowledgments}
This project was partly funded by the IDEX Universit{\'e} Grenoble Alpes innovation grant (AI4I-2018-2019) and the R{\'e}gion Auvergne-Rh\^{o}ne-Alpes (AISUA-2018-2019).

\bibliography{acl2019}

\begin{thebibliography}{30}
\expandafter\ifx\csname natexlab\endcsname\relax\def\natexlab#1{#1}\fi

\bibitem[{Agarwal et~al.(2018)Agarwal, Dymetman, and
  Gaussier}]{agarwal2018char2char}
Shubham Agarwal, Marc Dymetman, and Eric Gaussier. 2018.
\newblock Char2char generation with reranking for the e2e nlg challenge.
\newblock In \emph{Proceedings of INLG}, pages 451--456.

\bibitem[{Bahdanau et~al.(2014)Bahdanau, Cho, and Bengio}]{bahdanau2014neural}
Dzmitry Bahdanau, Kyunghyun Cho, and Yoshua Bengio. 2014.
\newblock Neural machine translation by jointly learning to align and
  translate.
\newblock In \emph{Proceedings of ICLR}.

\bibitem[{Bengio et~al.(2013)Bengio, L{\'e}onard, and Courville}]{BengioLC13}
Yoshua Bengio, Nicholas L{\'e}onard, and Aaron Courville. 2013.
\newblock Estimating or propagating gradients through stochastic neurons for
  conditional computation.
\newblock \emph{arXiv preprint arXiv:1308.3432}.

\bibitem[{Cheng et~al.(2016)Cheng, Xu, He, He, Wu, Sun, and
  Liu}]{cheng2016semi}
Yong Cheng, Wei Xu, Zhongjun He, Wei He, Hua Wu, Maosong Sun, and Yang Liu.
  2016.
\newblock Semi-supervised learning for neural machine translation.
\newblock In \emph{Proceedings of ACL}, pages 1965--1974.

\bibitem[{Chisholm et~al.(2017)Chisholm, Radford, and
  Hachey}]{chisholm2017learning}
Andrew Chisholm, Will Radford, and Ben Hachey. 2017.
\newblock Learning to generate one-sentence biographies from wikidata.
\newblock In \emph{Proceedings of EACL}, pages 633--642.

\bibitem[{Cho et~al.(2014)Cho, van Merrienboer, Gulcehre, Bahdanau, Bougares,
  Schwenk, and Bengio}]{cho2014learning}
Kyunghyun Cho, Bart van Merrienboer, Caglar Gulcehre, Dzmitry Bahdanau, Fethi
  Bougares, Holger Schwenk, and Yoshua Bengio. 2014.
\newblock Learning phrase representations using rnn encoder--decoder for
  statistical machine translation.
\newblock In \emph{Proceedings of EMNLP}, pages 1724--1734.

\bibitem[{Choi et~al.(2018)Choi, Yoo, and Lee}]{choi2018learning}
Jihun Choi, Kang~Min Yoo, and Sang-goo Lee. 2018.
\newblock Learning to compose task-specific tree structures.
\newblock In \emph{Proceedings of AAAI}.

\bibitem[{Du\v{s}ek and Jurc{\'{\i}}cek(2016)}]{Dusek2016}
Ond\v{r}ej Du\v{s}ek and Filip Jurc{\'{\i}}cek. 2016.
\newblock Sequence-to-sequence generation for spoken dialogue via deep syntax
  trees and strings.
\newblock In \emph{Proceedings of ACL}, pages 45--51.

\bibitem[{Gardent et~al.(2017)Gardent, Shimorina, Narayan, and
  Perez-Beltrachini}]{gardent2017creating}
Claire Gardent, Anastasia Shimorina, Shashi Narayan, and Laura
  Perez-Beltrachini. 2017.
\newblock Creating training corpora for micro-planners.
\newblock In \emph{Proceedings of ACL}.

\bibitem[{Gatt and Krahmer(2018)}]{Gatt2017}
Albert Gatt and Emiel Krahmer. 2018.
\newblock Survey of the state of the art in natural language generation: Core
  tasks, applications and evaluation.
\newblock \emph{Journal of AI Research}, pages 65--170.

\bibitem[{Gehrmann et~al.(2018)Gehrmann, Dai, Elder, and
  Rush}]{gehrmann2018end}
Sebastian Gehrmann, Falcon Dai, Henry Elder, and Alexander Rush. 2018.
\newblock End-to-end content and plan selection for data-to-text generation.
\newblock In \emph{Proceedings of INLG}, pages 46--56.

\bibitem[{Gu et~al.(2018)Gu, Im, and Li}]{gu2018neural}
Jiatao Gu, Daniel~Jiwoong Im, and Victor~OK Li. 2018.
\newblock Neural machine translation with gumbel-greedy decoding.
\newblock In \emph{Proceedings of AAAI}.

\bibitem[{He et~al.(2016)He, Xia, Qin, Wang, Yu, Liu, and Ma}]{he2016dual}
Di~He, Yingce Xia, Tao Qin, Liwei Wang, Nenghai Yu, Tie-Yan Liu, and Wei-Ying
  Ma. 2016.
\newblock Dual learning for machine translation.
\newblock In \emph{Proceedings of NIPS}, pages 820--828.

\bibitem[{Hochreiter and Schmidhuber(1997)}]{hochreiter1997long}
Sepp Hochreiter and J{\"u}rgen Schmidhuber. 1997.
\newblock Long short-term memory.
\newblock \emph{Neural computation}, 9:1735--1780.

\bibitem[{Jang et~al.(2017)Jang, Gu, and Poole}]{jang2016categorical}
Eric Jang, Shixiang Gu, and Ben Poole. 2017.
\newblock Categorical reparameterization with gumbel-softmax.
\newblock In \emph{Proceedings of ICLR}.

\bibitem[{Juraska et~al.(2018)Juraska, Karagiannis, Bowden, and
  Walker}]{JuraskaKBW18}
Juraj Juraska, Panagiotis Karagiannis, Kevin Bowden, and Marilyn~A. Walker.
  2018.
\newblock A deep ensemble model with slot alignment for sequence-to-sequence
  natural language generation.
\newblock In \emph{Proceedings of NAACL-HLT}, pages 152--162.

\bibitem[{Liu et~al.(2018)Liu, Yang, Wu, and Lee}]{liu2018improving}
Da-Rong Liu, Chi-Yu Yang, Szu-Lin Wu, and Hung-Yi Lee. 2018.
\newblock Improving unsupervised style transfer in end-to-end speech synthesis
  with end-to-end speech recognition.
\newblock In \emph{Proceedings of SLT}, pages 640--647.

\bibitem[{Luong et~al.(2015)Luong, Pham, and Manning}]{luong2015effective}
Thang Luong, Hieu Pham, and Christopher~D Manning. 2015.
\newblock Effective approaches to attention-based neural machine translation.
\newblock In \emph{Proceedings of EMNLP}, pages 1412--1421.

\bibitem[{Novikova et~al.(2017{\natexlab{a}})Novikova, Du{\v{s}}ek,
  Cercas~Curry, and Rieser}]{Novikova2017}
Jekaterina Novikova, Ond{\v{r}}ej Du{\v{s}}ek, Amanda Cercas~Curry, and Verena
  Rieser. 2017{\natexlab{a}}.
\newblock Why we need new evaluation metrics for nlg.
\newblock In \emph{Proceedings of EMNLP}, pages 2241--2252.

\bibitem[{Novikova et~al.(2017{\natexlab{b}})Novikova, Du\v{s}ek, and
  Rieser}]{NovikovaDR17}
Jekaterina Novikova, Ond\v{r}ej Du\v{s}ek, and Verena Rieser.
  2017{\natexlab{b}}.
\newblock The {E2E} dataset: New challenges for end-to-end generation.
\newblock In \emph{Proceedings of SIGDIAL}, pages 201--206.

\bibitem[{Qader et~al.(2018)Qader, Jneid, Portet, and
  Labb{\'e}}]{qader_hal-01950467}
Raheel Qader, Khoder Jneid, Fran{\c c}ois Portet, and Cyril Labb{\'e}. 2018.
\newblock {Generation of Company descriptions using concept-to-text and
  text-to-text deep models: dataset collection and systems evaluation}.
\newblock In \emph{Proceedings of INLG}.

\bibitem[{Ratcliff and Metzener(1988)}]{ratcliff1988pattern}
John~W Ratcliff and David~E Metzener. 1988.
\newblock Pattern-matching-the gestalt approach.
\newblock \emph{Dr Dobbs Journal}, pages 46--51.

\bibitem[{See et~al.(2017)See, Liu, and Manning}]{See2017}
Abigail See, Peter~J Liu, and Christopher~D Manning. 2017.
\newblock Get to the point: Summarization with pointer-generator networks.
\newblock In \emph{Proceedings of ACL}, pages 1073--1083.

\bibitem[{Sutskever et~al.(2014)Sutskever, Vinyals, and
  Le}]{sutskever2014sequence}
Ilya Sutskever, Oriol Vinyals, and Quoc~V Le. 2014.
\newblock Sequence to sequence learning with neural networks.
\newblock In \emph{Proceedings of NIPS}, pages 3104--3112.

\bibitem[{Tjandra et~al.(2017)Tjandra, Sakti, and
  Nakamura}]{tjandra2017listening}
Andros Tjandra, Sakriani Sakti, and Satoshi Nakamura. 2017.
\newblock Listening while speaking: Speech chain by deep learning.
\newblock In \emph{Proceedings of ASRU}, pages 301--308.

\bibitem[{Tjandra et~al.(2018)Tjandra, Sakti, and Nakamura}]{tjandra2018end}
Andros Tjandra, Sakriani Sakti, and Satoshi Nakamura. 2018.
\newblock End-to-end feedback loss in speech chain framework via
  straight-through estimator.
\newblock \emph{arXiv preprint arXiv:1810.13107}.

\bibitem[{Tu et~al.(2017)Tu, Liu, Shang, Liu, and Li}]{tu2017neural}
Zhaopeng Tu, Yang Liu, Lifeng Shang, Xiaohua Liu, and Hang Li. 2017.
\newblock Neural machine translation with reconstruction.
\newblock In \emph{Proceedings of AAAI}.

\bibitem[{Tu et~al.(2016)Tu, Lu, Liu, Liu, and Li}]{Tu2016}
Zhaopeng Tu, Zhengdong Lu, Yang Liu, Xiaohua Liu, and Hang Li. 2016.
\newblock Modeling coverage for neural machine translation.
\newblock In \emph{Proceedings of ACL}, pages 76--85.

\bibitem[{{Yang} et~al.(2017){Yang}, {Chen}, {Hakkani-Tür}, {Crook}, {Li},
  {Gao}, and {Deng}}]{Yang2017}
X.~{Yang}, Y.~{Chen}, D.~{Hakkani-Tür}, P.~{Crook}, X.~{Li}, J.~{Gao}, and
  L.~{Deng}. 2017.
\newblock End-to-end joint learning of natural language understanding and
  dialogue manager.
\newblock In \emph{Proceedings of ICASSP}, pages 5690--5694.

\bibitem[{Yin et~al.(2019)Yin, Lyu, Zhang, Osher, Qi, and
  Xin}]{yin2018understanding}
Penghang Yin, Jiancheng Lyu, Shuai Zhang, Stanley~J. Osher, Yingyong Qi, and
  Jack Xin. 2019.
\newblock Understanding straight-through estimator in training activation
  quantized neural nets.
\newblock In \emph{Proceedings of ICLR}.

\end{thebibliography}
\bibliographystyle{acl_natbib}

\end{document}